\documentclass[letterpaper]{article} % DO NOT CHANGE THIS
\usepackage{aaai25}  % DO NOT CHANGE THIS
\usepackage{times}  % DO NOT CHANGE THIS
\usepackage{helvet}  % DO NOT CHANGE THIS
\usepackage{courier}  % DO NOT CHANGE THIS
\usepackage[hyphens]{url}  % DO NOT CHANGE THIS
\usepackage{graphicx} % DO NOT CHANGE THIS
\def\UrlFont{\rm}  % DO NOT CHANGE THIS
\usepackage{natbib}  % DO NOT CHANGE THIS AND DO NOT ADD ANY OPTIONS TO IT
\usepackage{caption} % DO NOT CHANGE THIS AND DO NOT ADD ANY OPTIONS TO IT
\usepackage{algorithm}
\usepackage{algorithmic}

\usepackage{newfloat}
\usepackage{listings}
\DeclareCaptionStyle{ruled}{labelfont=normalfont,labelsep=colon,strut=off} % DO NOT CHANGE THIS
\floatstyle{ruled}
\newfloat{listing}{tb}{lst}{}
\floatname{listing}{Listing}
\title{A Survey of Theory of Mind in Large Language Models: Evaluations, Representations, and Safety Risks}

\author{
    Hieu Minh “Jord" Nguyen
    \thanks{Research done during the Apart Lab Fellowship}
}
\affiliations{
    Apart Research \\
    University of Science and Technology of Hanoi\\
    jordnguyen43@gmail.com
}

\begin{document}

\maketitle

\begin{abstract}

Theory of Mind (ToM), the ability to attribute mental states to others and predict their behaviour, is fundamental to social intelligence. In this paper, we survey studies evaluating behavioural and representational ToM in Large Language Models (LLMs), identify important safety risks from advanced LLM ToM capabilities, and suggest several research directions for effective evaluation and mitigation of these risks.

\end{abstract}

\section{Introduction}

Theory of Mind (ToM), first introduced in \citep{chimpanzeetom}, is the ability to attribute mental states to oneself and others. ToM is a fundamental aspect of human cognition and social intelligence, allowing inference or prediction of others' behaviours \cite{whatistom}.

Recent research has shown surprising ToM capabilities in LLMs. While results have been mixed on whether LLMs behaviourally exhibit robust ToM \cite{van-duijn-etal-2023-theory, ullman2023largelanguagemodelsfail}, research has also found that internal representations of self and others' belief states exist in current LLMs \cite{zhu2024language}. These representations have also been found to significantly affect ToM capabilities. 

Furthermore, the rapid developments of LLMs might cause significant ToM capability gains in the near future. This raises safety concerns in various contexts in user-facing applications and multi-agent systems, including risks such as privacy invasion and collective misalignment. 

In this paper, we survey studies evaluating behavioural and representational ToM, showing that: 1) LLMs can match humans performance on specific ToM tasks, 2) LLM ToM remains limited and non-robust, and 3) internal ToM representations suggest emerging cognitive capabilities. We then identify several safety implications from advanced LLM ToM in user-facing and multi-agent contexts. Finally, we recommend future research directions for better safety evaluation and risk mitigation.

\section{Empirical Landscape}

\subsection{Evaluating ToM in LLMs}

Recent work shows that performance of LLMs such as GPT-4 \cite{openai2024gpt4technicalreport} on ToM tests is comparable to 7-10 year-old children \cite{van-duijn-etal-2023-theory} or adult humans on some standard tasks like false belief or irony detection \cite{Strachan2024}. Some studies even demonstrate that LLMs can outperform humans on 6th-order ToM \cite{street2024llmsachieveadulthuman}. 

However, hard benchmarks designed specifically for LLMs such as BigToM \cite{gandhi2023understanding}, FANToM \cite{kim-etal-2023-fantom}, OpenToM \cite{xu-etal-2024-opentom}, Hi-ToM \cite{wu-etal-2023-hi}, and ToMBench \cite{chen-etal-2024-tombench} have most models stumped compared to humans on various ToM tasks. Furthermore, LLMs struggle with simple adversarial ToM examples, suggesting that current LLMs do not yet have fully robust ToM. \cite{shapira-etal-2024-clever, ullman2023largelanguagemodelsfail}

\subsection{Interpreting ToM in LLMs}

Meanwhile, work in interpreting LLMs has provided some evidence for genuine LLM ToM capabilities in the form of internal representations of others' beliefs. As \citet{zhu2024language} and \citet{bortoletto2024benchmarking} show, one can use linear probes \cite{alain2018understandingintermediatelayersusing} to extract from LLMs representations of belief states of others in ToM scenarios, and that steering LLMs with these probes can significantly affect performance in (false) beliefs identification questions. \citet{bortoletto2024benchmarking} also found that probing accuracy increasing with larger and fine-tuned LLMs, but that even small models like Pythia-70m can accurately represent beliefs from an omniscient perspective. Relatedly, \citet{jamali2023unveilingtheorymindlarge} demonstrate that specific neurons in deeper layers of LLMs closely correlate to ToM performance, paralleling neurons observed in human brains. 

This is further supported by research such as \cite{shai2024transformersrepresentbeliefstate}, who show that transformers can linearly represent data-generating processes in their residual stream and \cite{gurnee2024language}, who show that LLMs contains models of concepts such as space and time. This suggests that LLMs trained to predict text containing mental inference might also learn to represent mental states. 

%and (citation), who show that self-supervised models trained on sequential games such as Othello can learn a world model corresponding to board states.

% This suggests that similarly, LLMs trained to predict next tokens of text containing mental inference can develop capable ToM. Quoting Ilya Sutskever, "Predicting the next token well means that you understand the underlying reality that led to the creation of that token" (citation). We argue that by most definitions of ToM — having inner representations of others' beliefs and utilising these representations to identify and predict behaviours — LLMs do exhibit ToM, even with their current limitations on adversarial examples.

\subsection{Future Developments}

While current ToM capabilities in LLM remain nascent, future LLMs might prove more capable. Previous work shows that scaling \cite{van-duijn-etal-2023-theory} and prompting techniques \cite{wilf-etal-2024-think} can already substantially improve ToM performance. This trend seems likely to continue in the future \cite{bitterlesson}.

Moreover, developments in foundational LLM architectures \cite{gu2024mamba, peng-etal-2023-rwkv, liquidmodels}, test-time compute \cite{o1learningtoreason, snell2024scalingllmtesttimecompute}, and scaffolding \cite{davidson2023aicapabilitiessignificantlyimproved} should all be considered possible sources of capability gains in the near future.

\section{Safety Risks from Advanced ToM}

Improved LLM ToM could enable beneficial applications, such as improved simulations of human behaviour for social science \cite{park2023generativeagentsinteractivesimulacra, ziems-etal-2024-large}. However, similar to how humans might use ToM to better deceive or exploit others \cite{lyingtheoryofmind}, advanced ToM in LLMs is not without potential drawbacks.

Advanced ToM can be particularly concerning, as it both amplifies existing risks like privacy breaches and enables dangerous capabilities like sophisticated deception from misalignment. Moreover, some research has already demonstrated scenarios where improved LLM ToM could exacerbate potential harms. Therefore, safety risks from LLM ToM, both current and prospective, warrant serious consideration. We categorise these risks into two primary domains: \textbf{user-facing risks} and \textbf{multi-agent risks}.

\subsection{User-facing Risks}

\textbf{Privacy and social engineering}: \citet{staab2024beyond} found that LLMs are capable of accurately inferring demographic information of text authors, including age, gender, education level, and socioeconomic status, even when text anonymisation is applied. \citet{chen2024designingdashboardtransparencycontrol} successfully trained linear probes that achieve near-perfect accuracy in identifying internal representations of author characteristics, with 80\% average accuracy when transferred to real human conversations. Notably, the accuracy of these inferences improves as conversations progress.

With advanced ToM, these vulnerabilities might expand to more sensitive personal information, such as beliefs, preferences, and tendencies being extracted from seemingly innocuous conversations. This capability can worsen privacy invasion attacks, potentially allowing bad actors to launch more automated and personalised misinformation and social engineering campaigns \cite{privacysurveyYAO2024100211}.

\textbf{Deceptive behaviours}: Enhanced LLM ToM can enable more targeted and sophisticated deception across various scenarios, including fraud, misinformation, and model misalignment \cite{PARKdeception2024100988}. \citet{scheurer2024large} show that LLMs can strategically deceive their users when put under pressure, while \cite{järviniemi2024uncoveringdeceptivetendencieslanguage} and \cite{vanderweij2024aisandbagginglanguagemodels} demonstrate cases of LLMs misleading evaluators about their own capabilities. When humans are the targets of advanced LLM ToM, these risks become particularly pronounced. This challenge is made worse by the potential of misaligned LLMs deliberately lying about their own ToM capabilities during critical evaluations \cite{hubinger2021riskslearnedoptimizationadvanced, ngo2024the}.

\textbf{Unintentional anthropomorphism}: LLM ToM capabilities may lead to unintentional and misleading anthropomorphisation \cite{street2024llmtheorymindalignment}. ToM capabilities might be leveraged by an LLM or LLM developers to build unwarranted user trust, encourage emotional attachment, or exploit psychological vulnerabilities \cite{ELIZAEFFECT, kran2025darkbench}.

\subsection{Multi-agent Risks}

\textbf{Exploitation}: \cite{mukobi2023welfarediplomacybenchmarkinglanguage} found that some LLMs are highly exploitable in a variant of the zero-sum board game Diplomacy. If LLMs are capable of advanced ToM, they might attempt to exploit each other in interactions. \citet{perez-etal-2022-red} successfully used LLMs to red-team other LLMs, suggesting that in realistic scenarios, LLM agents could coax each other into unintended behaviours, such as misdirection, model control, or data extraction \cite{geiping2024coercing}.

\textbf{Catastrophic conflict escalation}: \citet{RiveraEscalation} demonstrated that many LLMs exhibit unpredictable patterns of catastrophic conflict escalation, sometimes leading to nuclear exchanges in LLMs multi-agent systems playing simulated war games. In realistic analogous scenarios, LLM agents with advanced ToM might escalate situations beyond human control. While \citet{wongkamjan-etal-2024-victories} shows that LLMs consistently outplay human players in Diplomacy, LLM-LLM communication remains limited due to their difficulty with deception and persuasion. More advanced ToM could increase effective conflict capabilities.

\textbf{Collective misalignment}: \citet{anwar2024foundational} argues that multi-agent alignment is not guaranteed by single-agent alignment. Advanced ToM can facilitate unwanted collusion between LLM agents. For instance, \citet{motwani2024secret} and \citet{mathew2024hidden} demonstrate that LLM agents can engage in information hiding during communications to secretly collude under supervision. This capability could significantly disrupt applications and safety frameworks involving multiple agents \cite{irving2018aisafetydebate,kenton2024on}.

\section{Future Research Directions}

While there are many LLM ToM benchmarks, most are limited to question-answering tasks and suffer from problems like data contamination and overfitting \cite{zhang2024carefulexaminationlargelanguage, alzahrani2024benchmarkstargetsrevealingsensitivity}.
To better address the aforementioned risks, we suggest that ToM evaluation frameworks should extend to more authentic LLM deployment scenarios, such as ToM in personal LLM assistants \cite{guan2023intelligentvirtualassistantsllmbased}, scaffolded multi-agent environments \cite{li2023theory}, or simulated social platforms \cite{tang2024gensimgeneralsocialsimulation}.
Additionally, several promising strategies exist that aim to retain useful ToM capabilities while mitigating safety risks in LLMs. Examples include model unlearning \cite{liu2024rethinkingmachineunlearninglarge, liWMDP}, activation/representation engineering \cite{turner2024steeringlanguagemodelsactivation, zouRepE}, and latent adversarial training \cite{casper2024defendingunforeseenfailuremodes}

\section{Conclusion}

We have surveyed behavioural and representational evaluations of LLM ToM and identified key risk cases from advanced ToM. As LLMs continue to advance, it is important and urgent that we develop robust evaluation frameworks and mitigation strategies to ensure the safe and beneficial development of ToM capabilities in AI systems.
\bibliography{aaai25}
\end{document}